\documentclass[10pt,twocolumn,letterpaper]{article}

\usepackage{cvpr}              

\usepackage{graphicx}
\usepackage{amsmath}
\usepackage{amssymb}
\usepackage{booktabs}

\usepackage[pagebackref,breaklinks,colorlinks]{hyperref}

\usepackage[capitalize]{cleveref}
\crefname{section}{Sec.}{Secs.}
\Crefname{section}{Section}{Sections}
\Crefname{table}{Table}{Tables}
\crefname{table}{Tab.}{Tabs.}

\newcommand{\figureref}[1]{Figure.~\ref{#1}}
\newcommand{\figref}[1]{Fig.~\ref{#1}}

\newcommand{\sref}[1]{Sec.~\ref{#1}}

\def\cvprPaperID{*****} 
\def\confName{CVPR}
\def\confYear{2022}

\begin{document}

\title{Bag of Tricks for Domain Adaptive Multi-Object Tracking}

\author{Minseok Seo\thanks{These authors contributed equally to this work.} \qquad Jeongwon Ryu\footnotemark[1] \qquad Kwangjin Yoon\thanks{Corresponding author.} \\
SI Analytics\\
70 Yuseong-daero 1689 beon-gil, Yuseong-gu, Daejeon 34047, Korea Rep.\\
{\tt\small \{minseok.seo; rjw0926; yoon28\}@si-analytics.ai}
}
\maketitle

\begin{abstract}
In this paper, \texttt{SIA\_Track} is presented which is developed by a research team from SI Analytics\footnote{\scriptsize{\url{https://si-analytics.ai/}}}. The proposed method was built from pre-existing detector and tracker under the tracking-by-detection paradigm. The tracker we used is an online tracker that merely links newly received detections with existing tracks. The core part of our method is training procedure of the object detector where synthetic and unlabeled real data were only used for training. To maximize the performance on real data, we first propose to use pseudo-labeling that generates imperfect labels for real data using a model trained with synthetic dataset. After that model soups scheme was applied to aggregate weights produced during iterative pseudo-labeling. Besides, cross-domain mixed sampling also helped to increase detection performance on real data. Our method, \texttt{SIA\_Track}\footnote{\scriptsize{\url{https://motchallenge.net/method/MOT=5537&chl=23}}}, takes the first place on MOTSynth2MOT17 track at BMTT 2022 challenge. The code is available on \small{\url{https://github.com/SIAnalytics/BMTT2022_SIA_track}}.
\end{abstract}

\section{Introduction}
\label{sec:intro}

Tracking multiple objects in a video is one of the most challenging problem in computer vision. This is because not only developing a powerful tracker is difficult, but also collecting and labeling data for training deep neural networks usually end up unsatisfactory due to their nature of data-hungry. Recent research called \textit{MOTSynth} \cite{fabbri2021motsynth} proposes to use synthetic dataset and shows that it can achieve state-of-the-art results on the MOT17 dataset \cite{milan2016mot16} by training recent methods \cite{bergmann2019tracking, zhou2020tracking} using solely synthetic data.

Inspired by their approach, we propose \texttt{SIA\_Track} of which detector is only trained with synthetic and unlabeled real data. We used MOTSynth as the synthetic dataset and training images of MOT17 as the unlabeled real data (ground-truth annotation of MOT17 was not used). The proposed method was built from pre-existing detector and tracker under the tracking-by-detection paradigm. We used YOLOX \cite{ge2021yolox} as a detector and associate detections using ByteTrack technique \cite{zhang2021bytetrack} which makes our method online tracker.

The core part of our method is training procedure of the object detector where synthetic and unlabeled real data were only used for training. To maximize the performance on real data, we first propose to use pseudo-labeling that generates imperfect labels for real data using a model trained with synthetic dataset. After that model soups \cite{wortsman2022model} scheme was applied to aggregate weights produced during iterative pseudo-labeling. Besides, cross-domain mixed sampling \cite{tranheden2021dacs} also helped to increase detection performance on real data. 

Our method, \texttt{SIA\_Track}, shows the comparable performance on MOT17 testset in terms of HOTA which is only $1.2\%$ less than the first ranker, StrongSORT \cite{du2022strongsort}. It is noteworthy that StrongSORT also used YOLOX detector like ours but they trained it with a whole training split of MOT17 rather than unlabeled one. Finally, the proposed method takes the first place on MOTSynth2MOT17 track at BMTT 2022 challenge.

\section{Method}
\label{sec:method}
In this section, we describe each component of \texttt{SIA\_Track} in detail.
In ~\sref{sec:baseline}, the starting point of  \texttt{SIA\_Track} is described. Then, we explain all methods that was applied to increase the performance of initial model in following sections.
~\sref{sec:st} describes the process of pseudo-labeling which generates imperfect labels for unlabeled target (real) dataset with a model trained with source (synthetic) data.
After that, a cross-domain mixed sampling method where source and pseudo-labeled target dataset were used for mosaic augmentation will be described in ~\sref{sec:dacs}, and finally, a model soup~\cite{wortsman2022model} method for improving domain generalization performance is described in ~\sref{sec:wa}.
\begin{figure*}[t]
    \centering
    \includegraphics[width=0.85\linewidth]{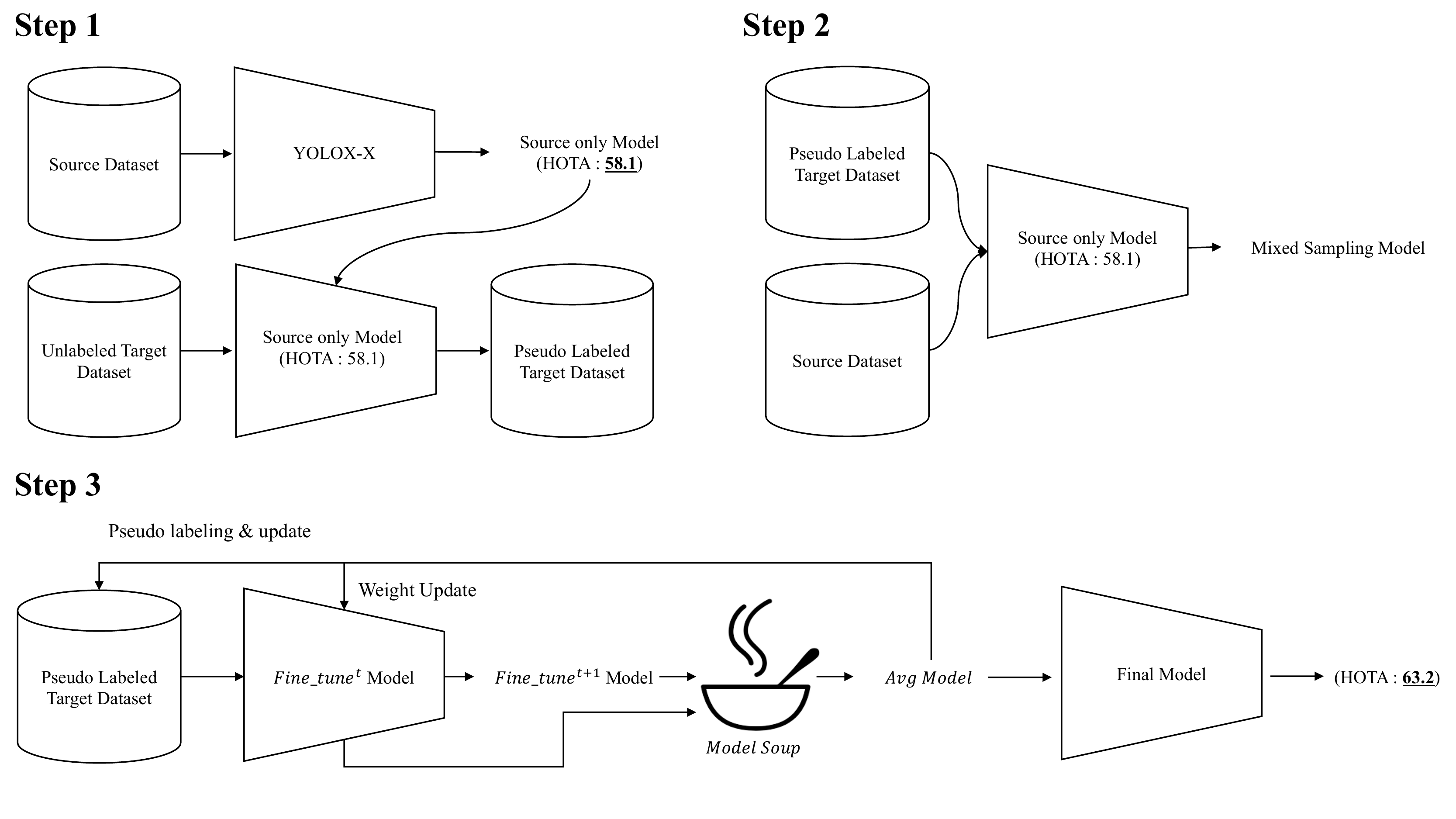}
    \caption{Overview of \texttt{SIA\_Track.} In \textbf{Step 1}, a pseudo-label of the unlabeled target dataset is generated with a warm-up model trained on the labeled source dataset. In \textbf{step 2}, the pseudo labeled target dataset and the labeled source dataset are trained through mixed sampling, and in \textbf{step 3}, model soup and iterative pseudo labeling are performed. }
    \label{fig:method}
\end{figure*}
\subsection{Overview}
\label{sec:over}
~\figureref{fig:method} is an overview of \texttt{SIA\_Track}.
The first training step of \texttt{SIA\_Track} is to generates a warm-up model $G_{1}$ trained only on the labeled source dataset $D^{S}$.
The warm-up model $G_{1}$ generates a pseudo labeled dataset $\tilde{D}^{T}$ for the unlabeled target dataset $D^{T}$.
After that, cross domain mixed sampling performs mosaic augmentations on $\tilde{D}^{T}$ and $D^{S}$ which consequently yields a model, $G_{2}$
The $G_{2}$ generated through the previous process is fine-tunned again after generating pseudo labels of the unlabeled target dataset (more accurate than the previous step).
The multiple-finetunned model generated through an iterative process is weight averaged through greedy soup to generate model $G_{t}$. (where $t$ is fine-tune stage)
\subsection{Warm-up model of \texttt{SIA\_Track}}
\label{sec:baseline}
We used ByteTrack~\cite{zhang2021bytetrack}, a recent leading approach in object tracking, as the beginning of our method.
In particular, YOLOX~\cite{ge2021yolox}, an object detector of ByteTrack, is a reasonable baseline for domain adaptive object detection because it includes domain generalization performance improvement schemes such as mosaic augmentation~\cite{bochkovskiy2020yolov4}, MixUp~\cite{zhang2017mixup} augmentation, and exponential moving average (EMA) weight update. 
We choose YOLOX-X since it shows the highest performance among YOLOX series.%
The YOLOX-X model trained solely on labeled source data achieved HOTA \textbf{58.1} on MOT17 testset.

\subsection{Pseudo Label Generation}
\label{sec:st}
Self-supervised training is a simple but effective method that is widely used in unsupervised domain adaptation~\cite{zhao2020collaborative, roychowdhury2019automatic, khodabandeh2019robust, saito2019semi}.
To apply the self-supervised training scheme, we generated pseudo labels for unlabeled target dataset with a model that achieved HOTA 58.1 in~\sref{sec:baseline}.
Since the low-confidence instances act as label noise~\cite{mei2020instance, zhang2021prototypical} and degrade the generalization performance of the model, instances with confidence less than 0.7 were removed.

\begin{figure}[t]
    \centering
    \includegraphics[width=0.8\linewidth]{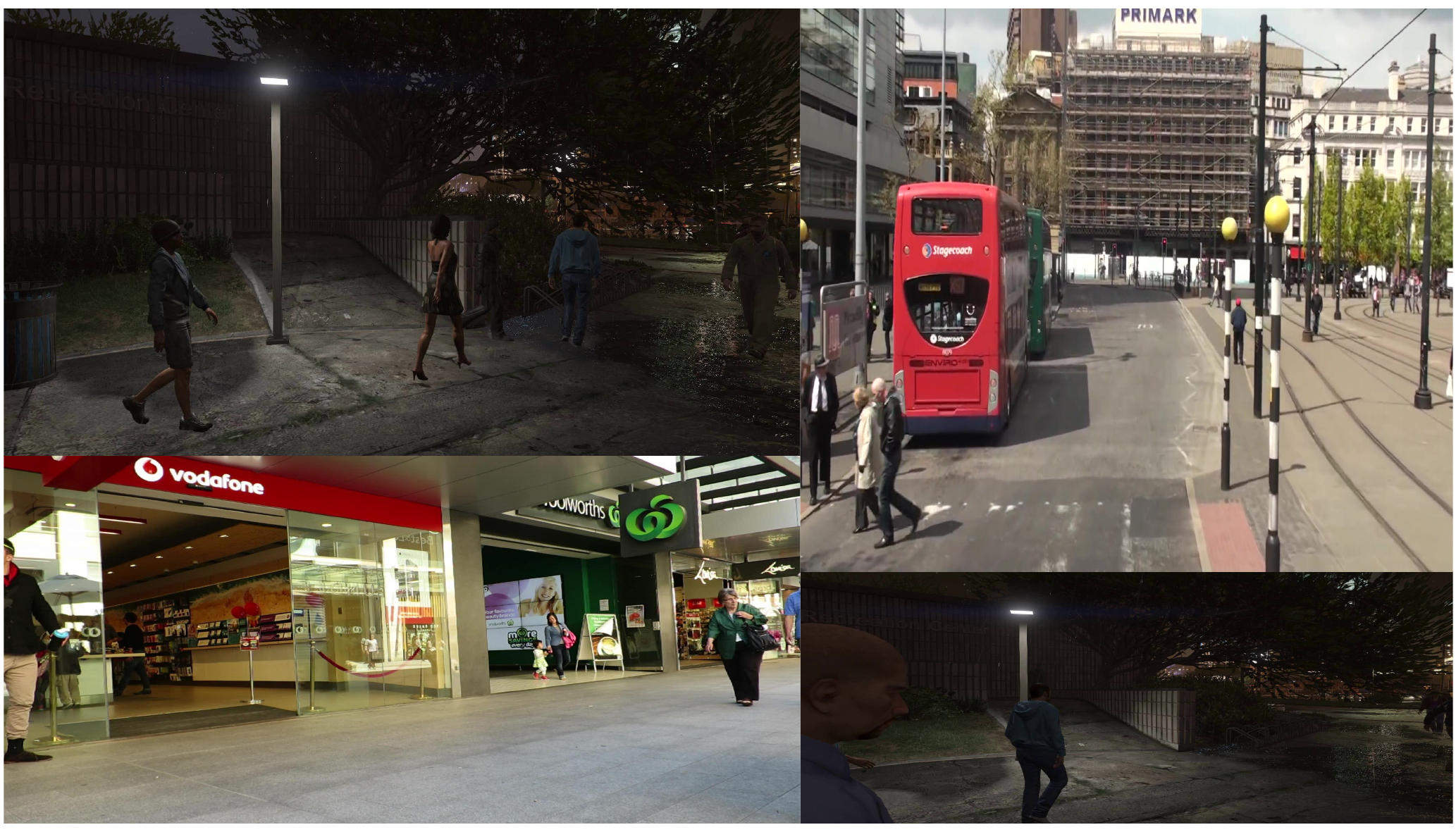}
    \caption{An example of mosaic augmentation of the target dataset and the source dataset.}
    \label{fig:mosaic}
\end{figure}

\setlength{\tabcolsep}{8pt}
\begin{table*}[t]
\begin{center}
\begin{tabular}{ccccccccc}
\hline
{\textbf{Rank}} & {\textbf{Method}} &   { {\textbf{HOTA} $(\uparrow)$}} &  {\textbf{MOTA} $(\uparrow)$} &  {\textbf{IDF1} $(\uparrow)$} &  {\textbf{AssA} $(\uparrow)$} &  {\textbf{DetA} $(\uparrow)$} &  {\textbf{FPS} $(\uparrow)$} \\
\hline \hline
 1st & \texttt{SIA\_Track} (ours)             & \textbf{63.2} & \textbf{77.6} & {76.8} & \textbf{62.3} & \textbf{64.4} & {24.2} \\ 
 2nd & \texttt{SDTracker} & {61.4} & {77.5} & \textbf{77.2} & 61.0 & 62.2 & {1.2} \\ 
 3rd & \texttt{PieTrack}  & {58.7} & 74.1 &	73.6 & 57.9	& 59.9 & {3.6} \\  
 4th & \texttt{fe\_Track} & {58.4} & 73.4 & 72.7 & 57.3	& 60.1 & {18.4} \\  
 5th & \texttt{Synth\_SORT} & {58.0} & 67.0 & 76.4 & 61.5 & 54.9 & {4.9} \\ 
 6th & \texttt{AIICMOT} & {56.1} & 73.6 & 68.1 & 53.1 & 59.6 & {19.7} \\ 
 7th & \texttt{FNT} & {46.3} & 55.7 & 60.2 & 46.3 & 46.6 & \textbf{39.5} \\ 
 8th & \texttt{Tracktor++} (Baseline) & {42.8} & 56.5 & 53.0 & 39.7 & 46.5 & {4.9} \\ 
\hline
\end{tabular}
\end{center}
\caption{
 MOTSynth-MOT leaderboard results.
}
\label{table:results}
\end{table*}

\subsection{Cross-domain Mixed Sampling}
\label{sec:dacs}
A naive approach for simultaneously training a pseudo labeled target dataset and a labeled source dataset is cross domain mixed sampling~\cite{tranheden2021dacs}.
A typical approach is to apply either cutmix or classmix to the source and target dataset.
However, mosaic augmentation and mixup are the already included in YOLOX-X, and they greatly affect the performance of YOLOX-X.
Therefore, as shown in ~\figref{fig:mosaic}, we applied mosaic augmentation using the source and target datasets, instead of cutmix and classmix.
The model trained through cross domain mixed sampling on the labeled source dataset and the pseudo labeled target dataset achieved HOTA \textbf{59.0} on MOT17 testset.

\subsection{Model soups}
\label{sec:wa}
\textit{Model soups}~\cite{wortsman2022model} is a domain generalization method that achieves state-of-the-art with a large gap in Imagenet-A and Imagenet-R datasets.
Model soups were used to prevent overfitting to the pseudo labeled target dataset (noisy label) and the labeled source dataset.
Therefore, the fine-tuned model and the non-fine-tuned model with pseudo-labeled target dataset are weighted averaged if the averaged model performed well on MOT17 trainset.

\subsection{Iterative Pseudo Labeling}
\label{sec:msst}
Iterative pseudo labeling is a method used for final performance tuning in domain adaptation competitions such as Visual Domain Adaptation (VisDA)~\cite{nam2019reducing} Challenge or unsupervised domain adaptation studies.
Iterative pseudo labeling improves performance because the pseudo label becomes more accurate as the number of iterations increases.
We finally achieved HOTA: \textbf{63.2} in the mot17 testset by applying model soup to the models generated through iterative pseudo labeling.
\section{Experiments}
\label{sec:experiments}
%
%
\subsection{Implementation Details}
\label{sec:imple}
\paragraph{Datasets} We used the MOTSynth dataset as the labeled source dataset according to the BMTT 2022 challenge rule, and the MOT17 testset as the evaluation dataset. We extracted frames from video using MOTSynth official code~\footnote{\url{https://github.com/dvl-tum/motsynth-baselines}}, and \texttt{train.json}, which consists of 103,320 images, was used for training.

\paragraph{Training} We trained a warm-up model using the ByteTrack official code\footnote{\url{https://github.com/ifzhang/ByteTrack}}. We trained it using 8 NVIDIA A100 GPUs, and the mini-batch size is 4. Note that, we trained 300 epochs because we cannot use pre-trained weights when training the warm-up model. The rest of the hyperparameters strictly follow the original code.

\paragraph{Fine-tuning} We changed the data augmentation type during fine-tuning according to the model soup recipe, and did not use EMA. During fine-tuning, a learning rate of 0.000025 was used, and all augmentation was implemented through albumentations. The applied augmentation types were \texttt{CoarseDropout, Downscale, ImageCompression, MultiplicativeNoise, MotionBlur, GaussNoise,} and \texttt{CLAHE}. For more detailed description and implementation, refer to the official code~\footnote{\url{https://albumentations.ai/}}.

\paragraph{Final Model} For tuning our final model, we added CBAM~\cite{woo2018cbam} and MixedStyle~\cite{zhou2021domain} in the step 2 and 3 model.

%



%
\subsection{Challenge Results}
\label{sec:result}

The performance of each entry at the MOTSynth2MOT17 track of the BMTT 2022 Challenge is measured by the HOTA metric\cite{luiten2021hota}. We evaluate our method on the MOT17 testset and get first place with 63.2 HOTA. The quantitative results are shown in Table~\ref{table:results}, along with MOTA\cite{bernardin2008evaluating} and IDF1\cite{ristani2016performance}. Specifically, our method not only achieves the highest accuracy except for IDF1 but also runs at numerically the second-highest running speed. However, the speed required by each method depends on the device being implemented.

%


%
%


%

\section{Conclusions}
\label{sec:conclusions}

In this challenge, we proposed a bag of tricks for domain adaptive multi-object tracking, which achieves 63.2 HOTA on the MOT17 dataset, ranking first on the leaderboard. However, there were problems in that the source dataset was excluded when performing Step 3 due to resource constraints, and the effective domain adaptation method in other synthetic datasets did not significantly affect the large-scale synthetic dataset (MOTSynth). Considering this, we plan to consider an effective domain adaptation method for large-scale synthetic data in the future.

{\small
\bibliographystyle{ieee_fullname}
\bibliography{sia}
}

\end{document}